\documentclass[10pt,twocolumn,letterpaper]{article}

\usepackage[pagenumbers]{wacv} 

\usepackage{graphicx}
\usepackage{amsmath}
\usepackage{amssymb}
\usepackage{booktabs}
\usepackage{times}
\usepackage{soul}
\usepackage{url}
\usepackage{adjustbox}
\usepackage[utf8]{inputenc}
\usepackage{algorithmic}
\usepackage[switch]{lineno}

\usepackage{epsfig}
\usepackage{amssymb}
\usepackage{multirow}
\usepackage{subcaption}

\DeclareUnicodeCharacter{0005}{}

\usepackage[dvipsnames]{xcolor}

\usepackage[pagebackref,breaklinks,colorlinks]{hyperref}

\usepackage[capitalize]{cleveref}
\crefname{section}{Sec.}{Secs.}
\Crefname{section}{Section}{Sections}
\Crefname{table}{Table}{Tables}
\crefname{table}{Tab.}{Tabs.}

\def\wacvPaperID{780} 
\def\confName{WACV}
\def\confYear{2025}

\title{Loose Social-Interaction Recognition in Real-world Therapy Scenarios}

\author{
    Abid Ali\textsuperscript{1 2} \hspace{1mm}
    Rui Dai\textsuperscript{4} \hspace{1mm}
    Ashish Marisetty\textsuperscript{1} \hspace{1mm}
    Guillaume Astruc\textsuperscript{1} \and
    Monique Thonnat\textsuperscript{1} \hspace{1mm}
    Jean-Marc Odobez\textsuperscript{3} \hspace{1mm}
    Susanne Th\"ummler\textsuperscript{1 2} \hspace{1mm}
    Francois Bremond\textsuperscript{1 2} \hspace{1mm}
    \\ \and
    \textsuperscript{1}INRIA \hspace{2mm} \textsuperscript{2}University Cote d'Azur \hspace{2mm}
    \textsuperscript{3}Idiap \hspace{2mm}
    \textsuperscript{4}Amazon Seattle
}
\begin{document}
\maketitle
\begin{abstract}
    The computer vision community has explored dyadic interactions for atomic actions such as pushing, carrying-object, etc. However, with the advancement in deep learning models, there is a need to explore more complex dyadic situations such as loose interactions. These are interactions where two people perform certain atomic activities to complete a global action irrespective of temporal synchronisation and physical engagement, like cooking-together for example. Analysing these types of dyadic-interactions has several useful applications in the medical domain for social-skills development and mental health diagnosis.
    
    To achieve this, we propose a novel dual-path architecture to capture the loose interaction between two individuals. Our model learns global abstract features from each stream via a CNNs backbone and fuses them using a new Global-Layer-Attention module based on a cross-attention strategy. We evaluate our model on real-world autism diagnoses such as our Loose-Interaction dataset, and the publicly available Autism dataset for loose interactions. Our network achieves baseline results on the Loose-Interaction and SOTA results on the Autism datasets. Moreover, we study different social interactions by experimenting on a publicly available dataset i.e. NTU-RGB+D (interactive classes from both NTU-60 and NTU-120). We have found that different interactions require different network designs. We also compare a slightly different version of our method (details in Section \ref{sec:temporal}) by incorporating time information to address tight interactions achieving SOTA results.
\end{abstract}

\section{Introduction} \label{intro}

Human activity recognition has been an active research area in the computer vision community for a wide range of applications, including health care, video surveillance, personality development, sports analytics, robotics, and so on.
In this domain, the analysis of human-human interaction and, more specifically, of dyadic interaction plays a central role. Dyadic interaction recognition has useful applications in the medical domain for social skills development, parent-child interaction therapy (PCIT), autism spectrum disorder (ASD) diagnosis, mental health diagnosis, education, etc. 


\begin{figure}[t]
\includegraphics[width=\linewidth]{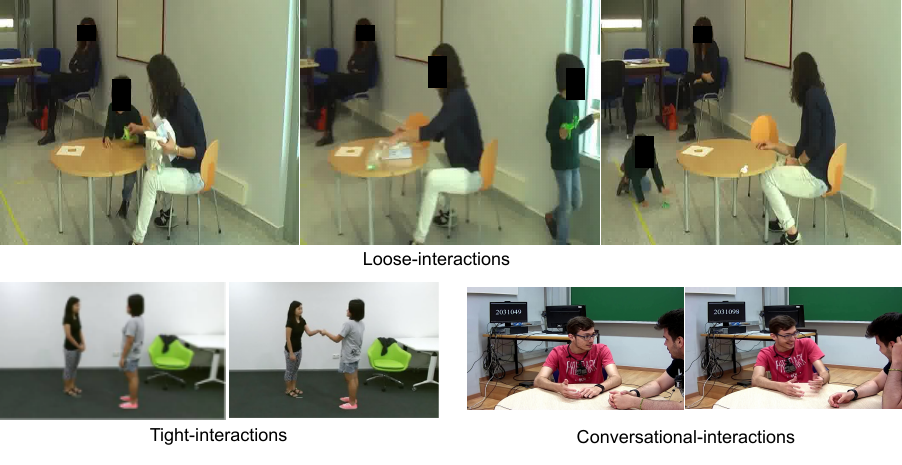}
\vspace*{-2em}
\caption{Different dyadic interaction types.} \label{interaction_types}
\end{figure}

Dyadic interactions can be categorised into three types i) \textit{tight interactions}, ii) \textit{conversational interactions}, and iii) \textit{loose interactions} as illustrated in Figure \ref{interaction_types}. Tight interactions are synchronised atomic actions with physical contact involved, such as shaking hands, hugging, etc. Moreover, tight interactive activities are composed of a few seconds with limited intra-class variance and high temporal synchronisation. They have been thoroughly studied in computer vision research. 
In the past decade, several high-performing Deep Convolutional Neural Networks (CNNs) models \cite{shafiqul2022hhi,khaire2022deep,zhu2021dyadic} have been designed to classify tight interactive actions with more than 90\% accuracy on lab-simulated public datasets such as SBU \cite{yun2012two}, ShakeFive \cite{van2016spatio}, NTU-RGB+D \cite{liu2019ntu}. 
Similarly, conversational interactions, such as engaging with one or more people during a meeting, a debate, or a talk, have been investigated in the literature for several purposes, including personality modelling \cite{agrawal2021multimodal,balazia2022bodily,palmero2022chalearn,curto2021dyadformer}.
A characteristic of such conversation-based interactions is that people have little mobility, and are primarily shot from front-facing cameras, and the main analysis goal is to detect and model the streams of conversational activities they exhibit like talking, eye-gaze aversion or contacts, facial expressions, and minimal hand gestures. 

In contrast, \textbf{loose interactions} are complex dyadic interactions, where two people individually perform a combination of asynchronous and asymmetric atomic actions that complete the global task without direct physical involvement. Loose dyadic interactions are long actions (more than one minute long) representing complex real-world scenarios such as \textbf{compound social interactions} (individuals performing different atomic actions within an activity), \textbf{spontaneous acting} (subjects acting freely without any specific guidance), \textbf{asynchronous and asymmetrical events} (both people performing independent and different atomic actions of their own, but together they complete an interactive activity such as celebrating a birthday), and \textbf{without physical engagement}. This type of activity usually consists of a leader and an assistant or helper.   
For example, in the activity of cooking, there is a leader (chef) who performs the main activity (cooking) by interacting in a loose manner with an assistant or helper (for instance, the assistant helps in chopping vegetables, providing required ingredients, etc.). 
The combination of such asynchronous and asymmetrical atomic actions generally has a global interaction for cooking activity. This type of weak interaction has not been explored much in the computer vision community to date. Therefore, there is a lack of research on recognising these complex activities that have loose asynchronous human-human interactions.

Additionally, in autism diagnosis, each ADOS \cite{lord2000autism} module corresponds to different tasks (actions of Loose-Interaction dataset) for severity evaluation. For example, the activity of \textit{imitation} responds to an analysis of child attention, gaze, and social skills. Classifying these loose interactions helps us to use each module for its appropriate task.


Existing deep learning models such as I3D \cite{carreira2017quo}, X3D \cite{feichtenhofer2020x3d}, and SlowFast \cite{feichtenhofer2019slowfast} etc. perform flawlessly on such tight interactive activities as they are atomic and temporally synchronised (kissing or hugging each other).
On the contrary, as discussed above, loose interactions are complex, having no such temporal synchronisation, neither are they symmetrical, and therefore are challenging for existing methods. Such interactions require a model that can exchange abstract-level information between the two individuals at different levels. This limits the capabilities of current models to be applied to such activities.

Furthermore, existing two-stream models, with early-fusion struggle to handle asynchronous and asymmetrical interactions (needs well-defined temporal synchronisation like the ones we see in tight interactions to perform well), while late-fusion models do not exchange sufficient information between the two streams. Therefore, mid-level global feature modelling is necessary to recognise loose interactions. 

Taking into account the above challenges, we propose a new architecture for loose-interaction recognition in social activities. The main contributions of this paper are as follows.
\begin{itemize}
    \item To our knowledge, we are the first to propose a new task of collaborative loose interactions, to focus on the recognition of asynchronous and asymmetrical loose social interactions in dyadic situations.  
    \item We propose a novel dual-path network for joint action recognition (composite social-interactive activities). The dual paths learn high-low-level multiscale visual features individually from two distinct inputs (leader and assistant) using \textbf{3D-CNNs}. The global abstract features are obtained through \textbf{Abstract Projection}. The action is recognised by performing a fusion via a novel \textbf{Global Layers Attention} (GLA) mechanism.
    
    \item We validate our method on a real-world dataset, depicting the loose social interactions of a clinician with a child during an autism diagnosis. Autism diagnosis data are recorded during the assessment of young children with ASD following the ADOS-2 protocols \cite{lord2000autism}. Besides this real-world dataset, we perform experiments on other datasets e.g., Autism, and NTU-RGB+D.
\end{itemize}

\section{Related Work}
\textbf{Video Classification}: CNNs have been very successful in learning 3D spatio-temporal representations for human activity recognition \cite{carreira2017quo}. Two-stream methods commonly used in combination of RGB and optical flow \cite{feichtenhofer2016convolutional}, with a special emphasis on video classification. SlowFast network \cite{feichtenhofer2019slowfast} has demonstrated the possibility of combining representations of different temporal resolutions (i.e. frame rates) to improve action recognition. Recently, with the advent of Transformers, several methods improved action recognition by incorporating attention. MViT \cite{fan2021multiscale} proposed pooling attention to learn spatio-temporal features at different scale. Video-Swin \cite{Liu_2022_CVPR} improved MViT using 3D shifted window modules for self-attention with patch merging after each spatial downsampling. 

Furthermore, Foundation models such as CLIP \cite{clip}, DINOV2 \cite{oquab2023dinov2}, and VideoMAE \cite{tong2022videomae} has been very useful for down-stream tasks such as action recognition, and action localisation \cite{park2023dual}. However, these methods are focused on general action recognition with little or no attention towards human-human interactions. Therefore, we adapt 3DCNN backbone to model pyramid of spatio-temporal features at higher spatial resolution (early layers) to low-level visual information (deeper layers) from two individual RGB inputs. On top of that we utilise cross-attention mechanism of transformers to build our interaction recognition architecture.

\textbf{Human Interaction Recognition} is a subdomain of recognition of actions. Lately, researchers combine CNN, RNN and GCN to recognise interactions from skeleton data \cite{perez2020interaction,ye2020human}. For instance, DR-GCN \cite{zhu2021dyadic} learns geometric and relative attention features from the two skeletons using their dyadic relational graph module. \cite{ito2022efficient} uses mid-fusion of 3-stream GCNs using inter and intra-body graphs to recognise interactions between two skeletons. Dyadformer \cite{curto2021dyadformer} proposed cross-subject layers using audio video inputs from two individuals to predict the personality of both individuals in long videos.
However, most of these methods are focused on dyadic short interactions with proper temporal synchronisation. A good skeleton input can effectively recognise short actions such as \textit{shaking hands} but could not model more complex loose interaction.
\begin{figure*}[t]
\centering
\includegraphics[width=0.98\linewidth]{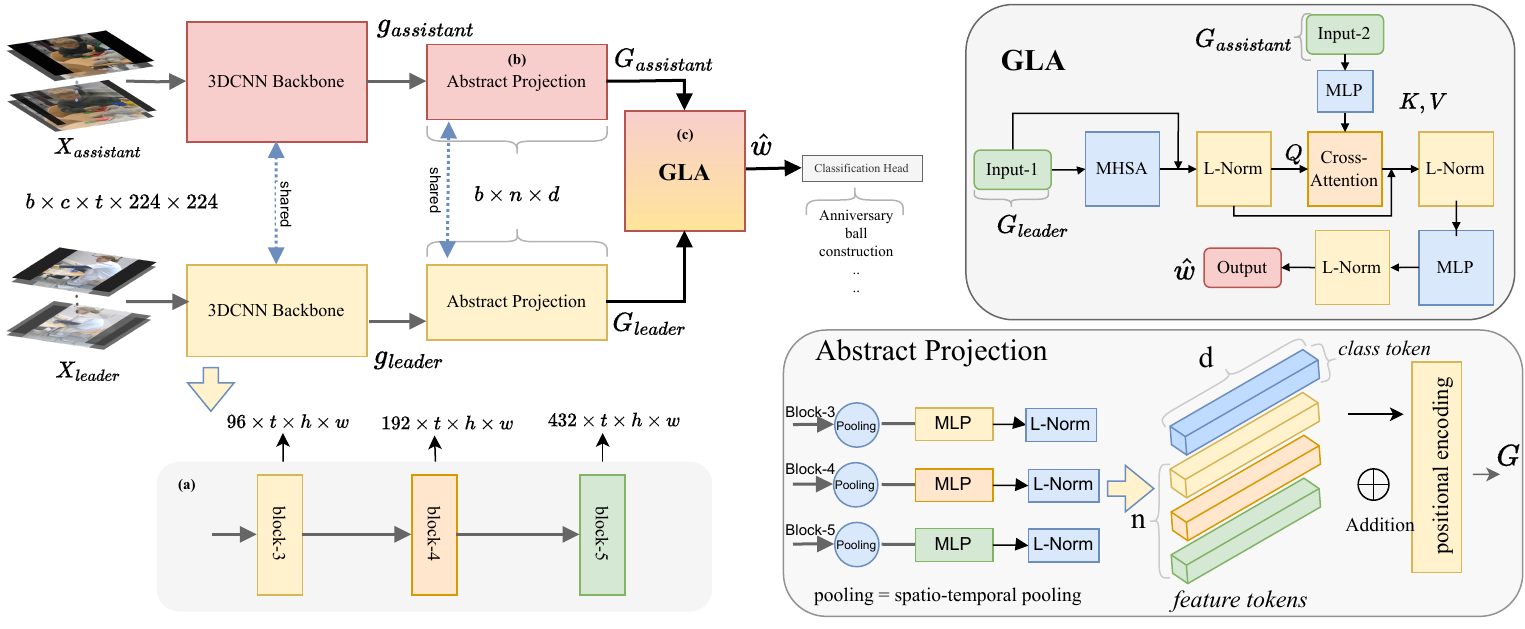}
\vspace*{-1em}
\caption{Our proposed architecture consists of (a) the Convolution backbone, (b) the Abstract Projection module, and (c) the GLA module. The model takes the input (leader and assistant) and outputs the action prediction score through the classification head.} \label{model}
\end{figure*} 

\textbf{ASD Recognition}: The Autism Diagnostic Observation Schedule (ADOS) \cite{lord2000autism}, a standard semi-structured test, was created by psychologists to identify ASD. The aim of ADOS, which can last up to two hours (four 30-minute sessions) and requires expert skills to carry out, is to assess the degree of social insufficiency in children. They designed individual modules to evaluate gaze, face gestures, body gestures and social-interactions of child during each session. In recent years, researchers have developed computer vision algorithms to address ASD behaviour recognition. Action recognition systems can use articulated posture structures, appearance, and motion information to study behavioural cues to address ASD diagnosis \cite{ali:hal-03447060,jeba_2022,Pandey_AP_Kohli_Pritchard_2020, negin2021vision}. \cite{ali:hal-03447060} uses two-stream I3D in a late-fusion manner to recognise autistic actions. Recently, \cite{Pandey_AP_Kohli_Pritchard_2020} proposed a guided weakly supervised method. They augment target autistic action classes with a general video dataset using posterior maximum likelihood for better behavioural posture learning. Unfortunately, these techniques focus primarily on repetitive or stemming behaviours of the child, ignoring the social complexity that occurs in interaction circumstances, which is a crucial component of  ASD diagnosis.
\section{Proposed Method} \label{x3d_arch}
In this section, we discuss our proposed architecture for recognising complex social activities that have asynchronous loose interactions. Our network addresses the key challenge: How to effectively model interactive activities that are asymmetrical and have no temporal synchronisation at the frame-level? To handle this situation, we came up with a dual path architecture. The network consists of four parts: i) \textbf{Convolution Backbone}, ii) \textbf{Abstract Projection Module}, iii) \textbf{Global Layers-Attention Module}, and iv) \textbf{Classification Head}, as shown in Figure \ref{model}. This is an end-to-end learning architecture. Each module is explained in the following.

\subsection{Terminology description}
Prior to modelling loose asynchronous interactions, we need to establish the roles of each individual involved in the interaction. Asynchronous loose interactions usually have a leader (focused on completing the whole action/task) and an assistant or helper (helping with minimal atomic actions). These roles could be reversed if the child is autistic (less socially active). Therefore, our input terms are defined as "Leader" and "Assistant" in the architecture.


\subsection{Convolution Backbone}
Our backbone learns the spatio-temporal multi-scale features of two distinct individual inputs in a dual-stream fashion. Both paths shares style design and parameters of convolution backbone reducing computation costs. We adapt a multi-scale 3D-CNNs having a pyramid of features with early layers operating at high spatio-temopral resolution modelling low-level visual features, and deeper layers at spatio-temporally coarse, but complex high-dimensional information. Let $X$ be the video snippet; then $X_{leader}^{b \times c \times t \times h \times w}$ and $X_{assistant}^{b \times c \times t \times h \times w}$ are the cropped images of the leader and assistant to extract $g^{b \times c \times t \times h \times w}$ coarse-fine features from the convolution backbone at different levels ($block_3$, $block_4$, and $block_5$). 

For each input $X_{leader}$, $X_{assistant}$ we extract the $g_{leader}$, and $g_{assistant}$, respectively. As both individuals perform these actions interactively (asynchronously and asymmetrically), the idea is to extract different symbolic spatio-temopral features in different blocks for fusion. As, the two individuals interact in a loose manner, fusing them locally (spatio-temporal level) does not benefit. Therefore, we need to compute global abstract information of each block for fusion, as shown in Figure \ref{model}(b). Experimentally, we find that utilising features from the last three blocks can efficiently recognise a complex activity (details in Section \ref{ablation}).


 


\subsection{Abstract Projection}\label{sec:projection}
Before concatenating, coarse-fine block-level features needs to be encoded to a common embeddings. First, we average-pool the spatio-temporal dimensions of each block to obtain a global context. As these loose actions are executed jointly, both individuals perform separate tasks to complete the entire activity regardless of time synchronisation at the frame-level: for instance, in the \textbf{Joint-game}\footnote{The activity is a part of ADOS assessment} activity, the clinician (leader) assembles different toys such as a dollhouse and plays with the child (assistant), where the child sets up a toy-lounge-furniture that involves \textit{pick and place small chair, and desk, etc.} and the clinician \textit{pick and move around a small doll}. Therefore, extracting a global context of the spatio-temporal features can better recognise these types of action. Second, we project these block-level features through MLP layer to learn abstract features of higher order at coarse-fine (blocks) level for each input as in Eq. \ref{eq_1}. We use GELU as activation function.
\begin{equation}\label{eq_1}
    L = MLP(activation(AvgPool3D(X)))
\end{equation}
The three projected layers are fused, obtaining $G^{b \times n \times d}$, where n represents the number of layers, which is three in our case, and d is the embedded feature vector of size 768. At this point, we concatenate a learnable classification token clfToken$^{b \times 1 \times 768}$ in dimension n along with the addition of a 1-D positional encoding {PE}$^{1\times 4 \times d}$ as shown in Eq. \ref{eq_2}. The same process is performed for $g_{leader}$ and $g_{assistant}$ individually.
\begin{equation}\label{eq_2}
    G^{b\times (n+1)\times d} = LNorm(fuse(L^{b\times d}*n, clf) + PE)
\end{equation}
where $G$ is the projected embedding of the blocks features for each stream, LNorm, L, n, clf, and PE represent layer normalisation \cite{ba2016layer}, embedded encoding, the number of layers, class-token, Positional Encoding, respectively. The abstract projection $G$ is executed separately for both paths $g_{leader}$, and $g{assistant}$, obtaining an embedding $G_{leader}$, and $G_{assistant}$  (as illustrated in Figure \ref{model}(b)).


\subsection{Global-Layer-Attention}
So far, the dual-paths independently learn abstract context and encode them through the abstract projection module as embeddings. Now, we need to integrate these embeddings from $G_{leader}$, and $G_{assistant}$ in such a way that we capture the asynchronous loose interaction between the two participants. To this end, we rely on an attention mechanism.
\begin{equation}\label{eq_3}
        Q = L\-Norm(G_{leader} + MHSA(G_{leader}))  
\end{equation}
\begin{equation}\label{eq_4}
        K = V = MLP(G_{assistant}))  
\end{equation}
In particular, let $Q$ be the query from $G_{leader}$ (in Eq. \ref{eq_3}) and the key and value $\{K, V\}$ (defined in Eq. \ref{eq_4}) from $G_{assistant}$ to perform a cross-attention in between the two inputs as in Eq. \ref{eq_5}. MHSA \cite{vaswani2017attention} stands for multi-head self-attention, MLP indicates multilayer perceptron. 

Our intuition is that the global high-level information (key-value pairs) of one person can be used to attend to the global context of the other person for better loose interaction recognition (as illustrated in Figure \ref{model}(c)) in an asynchronous and asymmetrical manner. We implement this intuition with the help of the proposed \textit{Global-Layer-Attention}. This module takes an input $G_{leader}$ and the corresponding vector from the interacted person $G_{assistant}$ and uses the multi-head cross attention mechanism to obtain the refined vector $\hat{w}^{b \times n+2 \times d}$ (as in Eq. \ref{eq_5}), where b, n, 2, d defines the batch-size, the number of layers, classification tokens, and the embedded dimensions (as illustrated in Figure \ref{model})(c). $T$ stands for transpose in Eq. \ref{eq_5}.
\begin{equation}\label{eq_5}
    \hat{w} = softmax(\frac{Q*K^T}{\sqrt{d_k}}) * V
\end{equation}
In our implementations, we use eight tokens, including two classification tokens and three abstract feature layers of each stream (obtained in Section \ref{sec:projection}) to perform cross attention between $G_{leader}$, and $G_{assistant}$. First, we apply a single MHSA  layer with eight attention heads in $G_{leader}$ followed by layer-normalisation to get $Q$. Before performing a cross-attention between $G_{leader}$ and $G_{assistant}$ we obtained $\{K, V\}$ by applying a single MLP layer on $G_{assistant}$. Finally, a cross-attention is performed between $Q$ and $\{K, V\}$. We use layer normalisation and add skip-connections at certain positions, as shown in Figure \ref{model}(c).


\subsection{Classification Head}
The output is taken by applying an MLP layer to the classification token of $\hat{Y}^{b \times 2 \times D}$ taken from the final output of the Global-Layer-Attention module, where $b$ is the batch size, 2 denotes the two classification tokens (one for each stream) and $D$ is the 768 embeddings.

\subsection{Temporal Synchronisation Modelling}\label{sec:temporal} 
To best accommodate temporally synchronised tight interactive actions, we made slight changes to our existing architecture. This slightly different design of our model captures frame-level temporal information in a synchronised and symmetrical manner. The main difference is, how we project the CNN layers, with specific changes in the Abstract Projection module. This variant of our model uses only the last CNN-block $block_5$ by pooling only spatial features, preserving the temporal information for both streams. The feature token is generated from $G^{b\times t\times d}$, where b is the batch-size, t represents the temporal frames (32 for NTU) and d are the embedded features as in Eq. \ref{eq_7}. The rest of the model is the same.
\begin{equation}\label{eq_7}
    G = MLP(Relu(AvgPool2D(block\_5\_layer)))
\end{equation}

\section{Experiments}
We have studied two types of interaction, i) loose interactions, and ii) tight interactions, and explored which modelling strategy (global abstract context, or temporal modelling) is effective for them.

In Section \ref{sec:loose} we experiment with the Loose-Interaction and the Autism \cite{Pandey_AP_Kohli_Pritchard_2020} datasets to study interactions in social therapy situations. We have found that global abstract features from the CNNs backbone can effectively address asymmetrical and temporally asynchronous interactions.


On top of that, we study the NTU-RGB+D \cite{liu2019ntu} dataset for tight interactions. Our study concludes that tight interactive actions are temporally synchronised and that a global abstract feature approach is not helpful. Tight interactions can be addressed with a slight change in the proposed architecture to model temporal information, as explained in Section \ref{sec:temporal}. 



\subsection{Loose-Interaction Dataset}
The Loose-Interaction dataset is actual children's assessment sessions recorded with clinicians at the hospital. 132-hour sessions were recorded following the ADOS-2 protocol to study the visual behaviour of children with the severity of autism.  Each child was diagnosed with a possible autism disorder during different interactive ADOS-2 activities. Long videos were classified into nine (9) interaction classes i.e., \textit{ anniversary, playing with bubbles, playing with ball, construction, demonstration, describing-image, imitation, joint-game, and puzzle}. Each action video is 2 - 4 minutes long, depending on the activity. Blurred, distorted, and out-of-frame videos were discarded, yielding a total of 845 trimmed videos, with 9 classes, captured with an HD camera at 30 fps. Some subjects did not perform the same activity; therefore, the dataset is not subject-oriented and highly imbalanced. In this paper, a total of 87 unique children's hour-long videos were used out of 132 videos. The statistics of the dataset are given in Table \ref{tab:dataset}
\begin{table}[h]
\begin{center}
    \begin{tabular}{lccc}
    \hline
    \textbf{Action}      & \textbf{\# of clips} & \textbf{\# of unique children} \\ \hline
    Anniversary          & 118                 & 63                            \\
    playing with bubbles & 110                 & 63                            \\
    playing with ball    & 67                  & 45                            \\
    construction         & 253                 & 38                            \\
    demonstration        & 45                  & 27                            \\
    describing image     & 43                  & 36                            \\
    imitation            & 121                 & 57                            \\
    joint game           & 41                  & 26                            \\
    puzzle               & 47                  & 30                            \\ \hline
    Total                & 845                 & 87(overlap exists)            \\ \hline
    \end{tabular}
\end{center}
\vspace{-0.7em}
\caption{Loose-Interaction dataset statistics.}
\label{tab:dataset}
\end{table}

We intend to release the dataset in multiple modalities after ethical approval. 
\subsection{Public Datasets}
Currently, there is no publicly available dataset for asynchronous loose interactions that fits our needs. Therefore, we evaluated our model on other closely related public datasets such as \textbf{Autism} \cite{Pandey_AP_Kohli_Pritchard_2020} and \textbf{NTU-RGB+D} \cite{liu2019ntu}. 

The \textbf{Autism} dataset was designed for the behavioural study of children with ASD under stress. It is more focused on the child and their autistic (repetitive) actions. It has some sort of loose interactions (\textit{in most videos, the clinician performs the activity with the child in a weak interactive manner}), but not asynchronous. Furthermore, the actions are fine-grained, short. There are 1333 clips with a total of 8 action classes. More description provided in the supplementary materials.

In the \textbf{NTU-RGB+D} \cite{liu2019ntu} dataset we consider only the interactive actions from both NTU-RGB+D 60 and 120 datasets achieving a total of 26 action classes in 13k video clips. Although actions fall into the category of tight-interactive activities, they could be useful for the validation of our model and for studying different methods that work for tight-interaction recognition. 

\begin{table}[h]
\begin{center}
\scalebox{0.85}{
\begin{tabular}{lccc}
\hline
Methods               & Input                         & \begin{tabular}[c]{@{}c@{}}Acc.\%\\ Mean\\\ \end{tabular} \\ \hline
2S-DRAGCN \cite{zhu2021dyadic}            & 2P skeleton                   & 33.84                                                    \\ \hline
GWSDR$_{rgb+flow}$ \cite{Pandey_AP_Kohli_Pritchard_2020}    & \multirow{3}{*}{scene}                         & 52.33                                                    \\
Mvit \cite{fan2021multiscale} &                               & 56.37                                                    \\ 
X3D \cite{feichtenhofer2020x3d}                 &                          & 63.50                                                    \\
DinoV2+TCN \cite{oquab2023dinov2} &                          & 63.98 \\
VideoMAE$_{finetuned}$ \cite{tong2022videomae} &                          & 64.50                                                                                                     \\ \hline
SlowFast \cite{feichtenhofer2019slowfast}           & \multirow{5}{*}{2P tracklets} & 20.06   \\
X3D$_{earlyfusion}$ \cite{feichtenhofer2020x3d}  &                               & 30.03                                                    \\
GWSDR$_{rgb}$ \cite{Pandey_AP_Kohli_Pritchard_2020} &                               & 41.43                                                    \\
CoarseFine \cite{Kahatapitiya_2021_CVPR}          &                               & 46.06                                                    \\
Mvit$_{latefusion}$ \cite{fan2021multiscale} &                               & 58.01                                                    \\ 
X3D$_{latefusion}$ \cite{feichtenhofer2020x3d}&                               & 64.01                                                   \\ 
DinoV2+TCN$_{latefusion}$ \cite{oquab2023dinov2} &                               & 65.86                                                   \\
VideoMAE$_{fine-tuned}$ \cite{tong2022videomae}&                               & 66.71                                                   \\ \hline
Proposed $\dag$       & \multirow{2}{*}{2P tracklets} & 37.03                                                   \\
\textbf{Proposed}              &                               & \textbf{72.04}                                                 \\ \hline
\end{tabular}
}
\end{center}
\vspace{-0.7em}
\caption{Baseline results on the Loose-Interaction dataset. $\dag$ defines our temporal model, 2P: means cropped tracklets of both persons, separately.} \label{tab:act}
\end{table}

\begin{table}[hb]
\begin{center}
\begin{tabular}{lcc}
\hline
Method                & Acc.\%         \\ \hline
ECO \cite{zolfaghari2018eco}                  & 61.4           \\
TSN \cite{wang2016temporal}                  & 68.0             \\
R(2+1)D \cite{ghadiyaram2019large}              & 69.8           \\
I3D \cite{carreira2017quo}                  & 69.3             \\
TSM \cite{lin1811temporal}                  & 69.8           \\
TSN+DR$_{rgb+flow}$ \cite{Pandey_AP_Kohli_Pritchard_2020}               & 70.1           \\
TSN+GWS+DR$_{rgb+flow}$ \cite{Pandey_AP_Kohli_Pritchard_2020}            & 72.5 \\
GWSDR$_{rgb+flow}$ \cite{Pandey_AP_Kohli_Pritchard_2020}           & 75.1           \\ \hline
\textbf{Proposed} & \textbf{76.3} \\
\textbf{Proposed}$_{rgb+flow}$ & \textbf{78.6} \\ \hline
\end{tabular}
\end{center}
\vspace{-0.7em}
\caption{Results and comparison with SOTA on the Autism dataset.}
\label{tab:autism}
\end{table}

\begin{table}[ht]
\begin{center}
\scalebox{0.85}{
\begin{tabular}{l|c|c|c}
\hline
Methods                & Input                 & Modality & \begin{tabular}[c]{@{}l@{}}Acc. \%\\ (CS)\end{tabular}                   \\ \hline
ST-LSTM \cite{perez2020interaction}              & \multirow{8}{*}{2P}   & \multirow{10}{*}{Skeleton} & 63.00  \\                                                
IRN$_{inter+intra}$ \cite{perez2020interaction}  &                        &  & 77.70                                                                              \\
GCA-LSTM \cite{perez2020interaction}               &                    &    & 73.00                                                                              \\
ST-GCN \cite{zhu2021dyadic}               &                 &       & 80.20                                                                              \\
AS-GCN \cite{zhu2021dyadic}               &                &        & 73.13                                                                              \\
2S-AGCN$_{uni\_joint}$  \cite{zhu2021dyadic}  &              &          & 83.19                                                                              \\
2S-AGCN$_{uni-bone}$ \cite{zhu2021dyadic}  &              &          & 85.25                                                                              \\
2S-DRAGCN \cite{zhu2021dyadic}             &             &           & 90.56                                                                              \\ \cline{1-2}
PoseC3D$_{limb}$ \cite{duan2022revisiting}     & \multirow{2}{*}{scene} &  & 94.91                             \\
PoseC3D$_{joint}$ \cite{duan2022revisiting}     &                  &      & 95.85                                                                              \\ \hline
I3D \cite{carreira2017quo}                  & \multirow{2}{*}{scene} & \multirow{5}{*}{RGB} &  82.00                                                          \\
Swin-Transformer \cite{Liu_2022_CVPR}    &                &        & 92.52                                                                              \\ \cline{1-2}
SlowFast \cite{feichtenhofer2019slowfast}            &          &        & 93.70                                                                              \\
Proposed               &        \multirow{2}{*}{2P}         &       & 95.02                                                                              \\
\textbf{Proposed $\dag$}  &             &           & \textbf{96.25}                                                                              \\ \hline
\end{tabular}
}
\end{center}
\vspace{-0.7em}
\caption{Results in the NTU-RGB+D dataset for interactive actions and comparison with the SOTA methods. $P1$ and $P2$ represent the results of using a single-person tracklet without interactions. $\dag$: means our temporal variant model. 1P (single), 2P (both) tracklets.}
\label{tab:ntu}
\end{table} 
\subsection{Experimental Details}
Complex loose interactions videos are temporally long and require higher temporal modelling (more than 60 frames) to capture the action completely. We have found that global abstract features from the CNNs backbone can effectively address such long complex (asymmetrical and temporally asynchronous) interactions. Utilising 3D-CNNs as our backbone has 2 main benefits. First, 3D-CNNs such as X3D \cite{feichtenhofer2020x3d} have longer temporal modelling capabilities due to multiscale temporal pooling compared to Transformers (Mvit, VideoSwin) \cite{fan2021multiscale,vaswani2017attention}. X3D can operate at 64 - 120 frames input with a low computational cost compared to VideoSwin. Second, training 3D-CNNs with a smaller dataset is comparatively better compared to Transformers without requiring additional training strategies. 

All networks were pre-trained on the kinetics-400 \cite{carreira2017quo} dataset. We kept the same training protocols for all experiments with a batch size of 8 and trained them for 100 epochs. For the proposed architecture, we have used the SGD optimiser with an initial learning rate of $0.003$ and a momentum of 0.9 at training time.

We use pre-processing explained in the supplementary materials to extract tracklets (individual bounding boxes) for the Loose-Interaction dataset. We used skeleton information for the extraction of tracklets in NTU-RGB+D, and Yolov5~\cite{glenn_jocher} with DeepSORT \cite{wojke_deepsort} for the Autism dataset. Additional details are provided in the Supplementary Materials. 
\subsection{Experiments on ASD datasets}\label{sec:loose}
\textbf{Experiments on Loose-Interactions:} we first evaluate our proposed dual-path architecture on the Loose-Interaction dataset, the results are reported in Table \ref{tab:act}. Handling the long temporal duration of the Loose-Interaction dataset, we perform several experiments with different temporal sizes, including 32, 64, 80, and 120 frames. The best results were achieved with the 80-frame snippet. All baseline models follow the same protocols. For a fair comparison, an additional TCN layer is used for longer temporal modelling in transformer-based baselines such as Mvit \cite{fan2021multiscale} and VideoMAE \cite{tong2022videomae}. 

We compare our proposed method with existing 3D-CNNs, GCN, and Transformer-based architectures. 

GWSDR \cite{Pandey_AP_Kohli_Pritchard_2020} uses an additional optical-flow modality to capture motion. Interestingly, this method did not perform well on Loose-Interaction dataset. We notice, their model greatly benefits from a weak co-learning strategy by training on other similar atomic actions found in large datasets (Kinetics). However,
the loose-Interaction dataset has unique and composite actions, different from the Kinetics action classes. Thus, pretraining in such a manner does not fully help the model to converge. Secondly, the actions are longer, asymmetrical, and temporally asynchronous for this method to capture well. 2S-DRAGCN \cite{zhu2021dyadic} uses skeletons for dyadic-interactions. However, the model is designed to better capture synchronised atomic interactions (as in the case of NTU-RGB+D) compared to complex actions in Loose-Interaction dataset.

We further investigate SOTA transformers-based action classification architectures for loose-interaction tasks. We experiment with MViT \cite{Li_2022_CVPR} (multiscale design), VideoMAE \cite{tong2022videomae} (general mask learning model) and DINOV2 \cite{oquab2023dinov2} (foundation model). We use MViT-small with an additional TCN layer for longer temporal pooling. However, the model did not converge due to the size of the dataset. Next, we use fine-tuning strategies by freezing a few layers at a time to let the model learn and converge for each input separately. Later, we use a latefusion strategy as shown in Table \ref{tab:act}. We experiment with the same strategy for VideoMAE \cite{tong2022videomae}. A common problem with these methods is the small size of the dataset and higher computational costs. Furthermore, they works well on atomic actions compared to long complex activities. Also, the architectures perform poorly if trained separately, reducing computational costs, but could not converge fully if both inputs are jointly trained.  

Lastly, we use a multilayer TCN model for temporal modelling of spatial features extracted from DinoV2 \cite{oquab2023dinov2} and MLP layers for classification. DINOV2+TCN performs well with only temporal modelling but is unable to capture more complex actions due to no spatial interaction learning.

\textbf{Experiments on Autism:} in this section, we perform experiments on the Autism dataset where we use 32 frames as input snippets. Our proposed network achieves new SOTA results (Table \ref{tab:autism}) in the Autism dataset with an increase in accuracy of 1.2\%. The existing methods in Table \ref{tab:autism} use full-frame in their experiments. GWSDR \cite{Pandey_AP_Kohli_Pritchard_2020} utilises an additional optical flow stream in conjunction with the rgb scene to capture atomic actions in this dataset. Furthermore, they greatly benefited from other large-scale datasets using their guided weak supervision technique. However, our proposed method achieves higher accuracy by just using rgb tracklets of the two individuals. To make a fair comparison with them, we further use an additional optical flow stream for this experiment in latefusion manner, achieving an additional 2.3\% increase in accuracy.

\begin{table}[tb]
\centering
\begin{tabular}{ll}
\hline
Input                                   & Acc. \% \\ \hline
MHSA(G$_{assistant}$), MLP(G$_{leader}$) & 71.3    \\
MLP(G$_{assistant}$), MHSA(G$_{leader}$) & 72.0    \\ \hline
\end{tabular}
\vspace{-0.2em}
\caption{Impact of swapping the inputs of the GLA module.}
\label{tab:input_swap}
\end{table}

\begin{table}[h]
    \begin{center}
    \begin{tabular}{c|c}
            \hline
            \textbf{Network streams} & \textbf{Acc.\%} \\ \hline
            leader            & 55.62            \\
            assistant        & 51.84            \\
            leader+assistant$_{LF}$       & 64.02            \\
            \textbf{Proposed}     & \textbf{72.04}   \\ \hline
    \end{tabular}
    \end{center}
    \vspace{-.8em}
    \caption{Importance of each stream in our design choice. LF: stands for late-fusion}
    \label{tab:stream}
\end{table}

\begin{table}[h]
    \begin{center}
    \begin{tabular}{c|c}
 \hline
        \textbf{Layer Selection}              & \textbf{Acc.\%} \\ \hline
        $conv_5$                      & 60.0     \\
        All ($block_1$ -- $block_5$)     & 48.0     \\
        $block_2, block_3, block_4, block_5$ & 66.0     \\
        \textbf{$block_3, block_4, block_5$}        & \textbf{72.0}     \\ \hline
    \end{tabular}
    \end{center}
    \vspace{-.8em}
    \caption{Experimenting with the number of layers to model abstract features.}
    \label{tab:cnn_layer}
\end{table}

\begin{table}[h]
\begin{center}
\begin{tabular}{l|llc|l}
\cline{1-2} \cline{4-5}
Attention       & Acc.\% &  & GLA modules & Acc.\% \\ \cline{1-2} \cline{4-5} 
Self-attention  & 38.5   &  & 2           & 44.2   \\
Cross-Attention & 72.0   &  & 1           & 72.0   \\ \cline{1-2} \cline{4-5} 
\end{tabular}
\caption{Ablation study on attention mechanism and analysing the number of GLA modules used.}
\label{tab:attn-tf_layer}
\end{center}
\end{table}

\begin{table}[h]
    \begin{center}
    \begin{tabular}{c|c}
            \hline
            \textbf{Components}                                                                   & \textbf{Acc.\%}                       \\ \hline
            W/o AP, W/o GLA                                                            & 64.0  \\
            W AP, W/o GLA & 66.5     \\
            \textbf{W AP, W GLA}                                               & \textbf{72.0} \\ \hline
    \end{tabular}
    \end{center} 
    \vspace{-0.8em}
    \caption{Experiments with and without the Abstract Projection (AP), and GLA module. \textbf{W} and \textbf{W/o} means \textbf{with} and \textbf{without}, respectively.}
    \label{tab:components}
\end{table}

\subsection{Experiments on tight interactions}\label{sec:tight}
On the NTU-RGB+D dataset, we only used interactive action classes. For this reason, we compare our method with the SOTA methods in interactive action classes, as shown in Table \ref{tab:ntu}. Our main network achieves results comparable to those of the SOTA. We further extend our network to model temporal synchronisation. Our slightly changed network achieves SOTA results in the NTU-RGB+D dataset as reported in Table \ref{tab:ntu}. Our original proposed method is also comparable to that of SOTA.  The validation of these two designs on the loose and tight interactive datasets shows the usability of different network design strategies for different synchronisations and symmetry.

\section{Ablation Study} 
\label{ablation}
Here, we discuss several ablation experiments that validate our design choices, using the Loose-Interaction dataset as it is the most relevant one. 
More specifically, we analyse
i) the influence of the network inputs swapping; 
ii) the CNNs embedding layers;
iii) the attention type; 
iv) the number of GLA modules used for attention; 
v) the importance of each component used.

To evaluate the impact of leader and assistant on model performance, we swapped the input of the GLA module. Table~\ref{tab:input_swap} validates that there is an asymmetric and asynchronous behaviour, where the leader is interested in completing the activity by having a loose interaction with the assistant. The small drop in accuracy we observe is due to the fact that when the child is autistic, they are not fully involved in the activity, as shown in Table \ref{tab:stream}. Thus, the information he/she carries is smaller compared to that of their partner. This validates our proposed way of processing the leader and assistant streams as in Figure \ref{model}(c).

Furthermore, to understand the usefulness of each input, we experiment with training the $X3D$ \cite{feichtenhofer2020x3d} model separately for both the leader and the assistant. The results in Table \ref{tab:stream} show that the leader stream is more accurate in recognising activity than the assistant. Combining both streams using our proposed GLA module can greatly improve the prediction of such loose interactions. In addition, we provide more ablation study about the number of attention heads used in GLA module and depth of backbone in the Supplementary Materials Section \textcolor{Red}{4.1}. 

\begin{table}[h]
\begin{center}
\begin{tabular}{ccccc}
\hline
Device & \begin{tabular}[c]{@{}c@{}}Flops\\ (G)\end{tabular} & \begin{tabular}[c]{@{}c@{}}Param\\ (M)\end{tabular} & \begin{tabular}[c]{@{}c@{}}Infer. Time\\ Data (ms)\end{tabular} & \begin{tabular}[c]{@{}c@{}}Infer. Time\\ Model (ms)\end{tabular} \\ \hline
CPU    & -                                               & -                                              & 1760                                                               & 3830                                                                \\
GPU    & 76.99                                               & 10.54                                               & 1560                                                               & 80.90                                                               \\ \hline
\end{tabular}
\caption{(Computational complexity analysis. We use a 10 clips testing strategy. We show computational complexity in GFLOPs for a single clip input and inference time of the model and data in milliseconds. Provided inference time is on both CPU (Xeon Silver 4215) and Tesla v100 GPU for a single batch size. Infer. is short for Inference)}
\label{tab:complexity}
\end{center}
\end{table}

Next, we analyse the pyramid of high- and low-level features extracted from the 3D-CNN backbone for further projection. Specifically, features of $block_3$, $block_4$, and $block_5$ of the 3D-CNNs backbone. We compare this design choice with other experimental approaches from using only the $block_5$ block to utilising all blocks, $block_1 - block_5$, from both streams. The results of these experiments are given in Table \ref{tab:cnn_layer}. This analysis demonstrates that earlier low-level features are not as important compared to deep-layer features. 

Moreover, the proposed method is based on cross-attention strategy to perform attention using the novel GLA module. A cross-attention between two different input streams can emphasise the correlation between them efficiently. We compared this design choice with self-attention to validate its importance. Table \ref{tab:attn-tf_layer} defines this comparison between different attention approaches. Next, our network has only one GLA module for this fusion between the two paths using attention. However, we have noticed a drop in the efficiency of our network when increasing the complexity of the model (increasing the number of GLA modules). One possible reason for this is the small size of the dataset to fully use a more dense architecture. This analysis is reported in Table \ref{tab:attn-tf_layer}. Additionally, we have evaluated the usefulness of the Abstract Projection module and the GLA module by experimenting without each of them. Results are described in Table \ref{tab:components}.

\subsection{Computational Complexity Analysis} 
We evaluate our model's computational complexity using GLOPs and inference time for a single input to validate its use for devices with limited resources in real-time, as shown in Table \ref{tab:complexity}. Our model is light-weight as the backbone and abstract projection modules are shared between the two paths, thus our model uses only 10.54M parameters having 76.99 GFLOPs. Furthermore, it takes only 3.8 seconds to run a single input video on the CPU (Xeon Silver 4215) and 1.76 seconds to process a single batch input. Similarly, with a single Tesla v100 GPU our model takes 0.80 seconds to run a single input video. This validates our model is efficient in terms of computational complexity and can work on resource-constrained devices in real-time. 
\section{Conclusion and Future Work}
Recognising complex social-interaction between two individuals performing an action is a challenging task. We propose a new direction for human-human action recognition having loose interactions. To address this challenging task, we design a new architecture that attends to temporally unsynchronised loose-interactive actions using our novel Global-Layer-Attention module. We validate our network in social therapy scenarios for Loose-Interactions and Autism datasets. Our model achieves SOTA results on these two datasets. To demonstrate our network generalisability in tight-interactive actions, we experiment with the NTU-RGB+D dataset. Our model achieves higher results by slightly changing the model design to capture synchronised interactions. Our proposed method has certain limitations to handle all types of social interactions. One possible solution would be to design an adaptive temporal synchronisation module that can model symmetrical and asymmetrical time synchronisation between two people. 

In the future, our next goal is to detect the activities of autistic children in untrimmed videos. With the help of action detection and recognition methods, the final goal will be to generate severity reports for autistic children by automating the autism diagnosis process.

{\small
\bibliographystyle{ieee_fullname}
\bibliography{egbib}
}
\end{document}